\def\year{2019}\relax
\documentclass[letterpaper]{article} 
\usepackage{aaai19}  
\usepackage{times}  
\usepackage{helvet}  
\usepackage{courier}  
\usepackage{url}  
\usepackage{graphicx}  
\usepackage{comment}
\usepackage{color}
\usepackage{multirow}
\usepackage{amsmath}
\usepackage{subfig}

\usepackage[utf8]{inputenc} 
\usepackage[T1]{fontenc}    
\usepackage{url}            
\usepackage{booktabs}       
\usepackage{amsfonts}       
\usepackage{nicefrac}       
\usepackage{microtype}      
\usepackage{latexsym}
\usepackage{kotex}
\usepackage{verbatim}

\newcommand\sh[1]{\textcolor{black}{#1}}
\newcommand\shh[1]{\textcolor{black}{#1}}

\newcommand\nj[1]{\textcolor{black}{#1}}
\newcommand{\etal}{\textit{et al}. }

\begin{document}
%
\title{Semantic Sentence Matching with Densely-connected\\ Recurrent and Co-attentive Information
}
\author{  
Seonhoon Kim$^{1,2}$, Inho Kang$^1$, Nojun Kwak$^2$  \\
$^1$Naver Search, $^2$Seoul National University\\
\\
\texttt{ \{seonhoon.kim|once.ihkang\}@navercorp.com, nojunk@snu.ac.kr} \\
}
\maketitle

\begin{abstract}
Sentence matching is widely used in various natural language tasks such as natural language inference, paraphrase identification, and question answering. For these tasks, understanding logical and semantic relationship between two sentences is required but it is yet challenging. Although attention mechanism is useful to capture the semantic relationship and to properly align the elements of two sentences, previous methods of attention mechanism simply use a summation operation which does not retain original features enough. Inspired by DenseNet, a densely connected convolutional network, we propose a \textit{densely-connected co-attentive recurrent neural network}, each layer of which uses concatenated information of attentive features as well as hidden features of all the preceding recurrent layers. It enables preserving the original and the co-attentive feature information from the bottommost word embedding layer to the uppermost recurrent layer. To alleviate the problem of an ever-increasing size of feature vectors due to dense concatenation operations, we also propose to use an autoencoder after dense concatenation. We evaluate our proposed architecture on highly competitive benchmark datasets related to sentence matching. Experimental results show that our architecture, which retains recurrent and attentive features, achieves state-of-the-art performances for 
\shh{most of} the tasks.
\end{abstract}

\section{Introduction}

Semantic sentence matching, a fundamental technology in natural language processing, 
requires lexical and compositional semantics.
In paraphrase identification, sentence matching is utilized to identify whether two sentences have identical meaning or not. In natural language inference also known as recognizing textual entailment, it determines whether a hypothesis sentence can reasonably be inferred from a given premise sentence. In question answering, sentence matching is required to determine the degree of matching 1) between a query and a question for question retrieval, and 2) between a question and an answer for answer selection. However identifying logical and semantic relationship between two sentences is not trivial due to the problem of the semantic gap \cite{liu2016deep}.

\begin{figure*}
  \centering
  \includegraphics[width=0.9\linewidth]{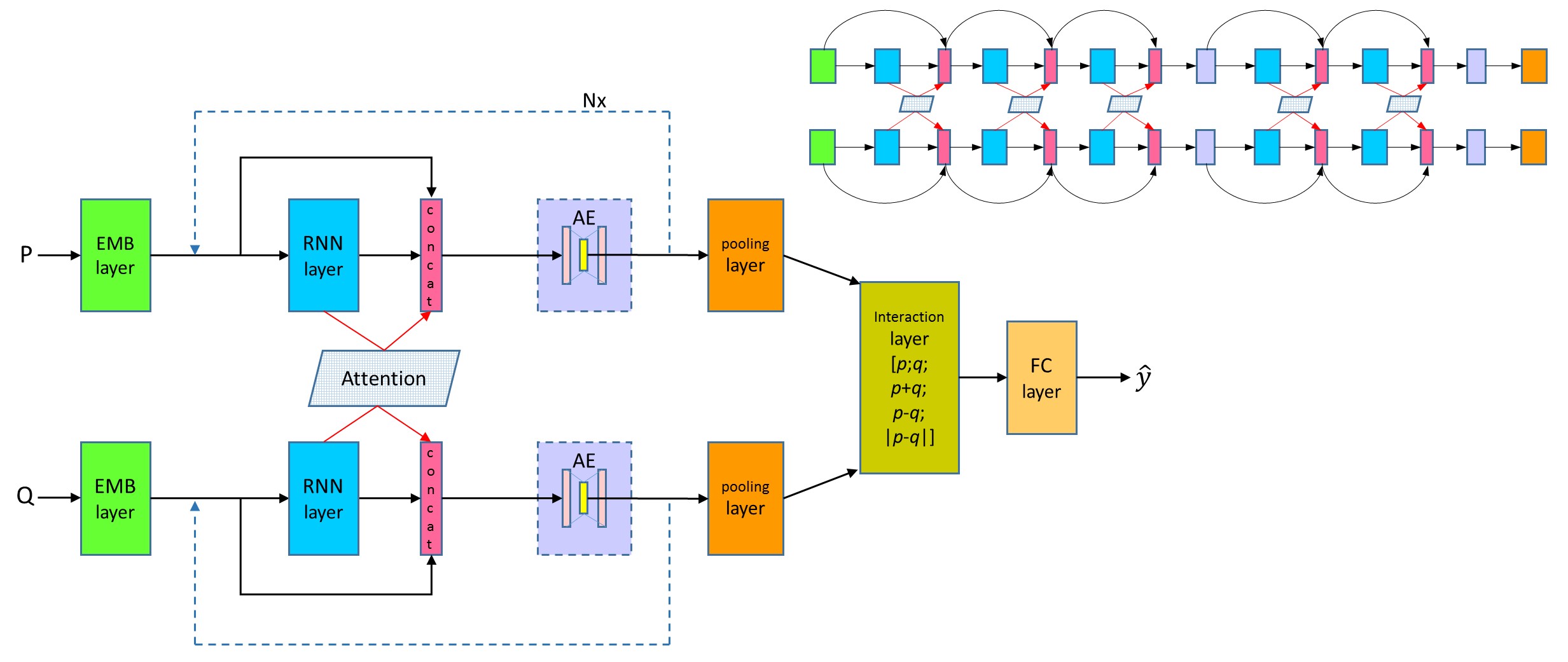}
  \caption{General architecture of our Densely-connected Recurrent and Co-attentive neural Network (DRCN). Dashed arrows indicate that a group of RNN-layer, concatenation and AE can be repeated multiple ($N$) times (like a repeat mark in a music score). The bottleneck component denoted as AE, inserted to prevent the ever-growing size of a feature vector, is optional for each repetition. The upper right diagram is our specific architecture for experiments with 5 RNN layers ($N=4$).}  
  \label{fig:arch}
\end{figure*}

Recent advances of deep neural network enable to learn textual semantics for sentence matching. Large amount of annotated data such as Quora~\cite{quora}, SNLI~\cite{snliemnlp2015}, and MultiNLI~\cite{williams2017broad} have contributed significantly to learning semantics as well. In the conventional methods, a matching model can be trained in two different ways \cite{gong2018natural}. The first methods are sentence-encoding-based ones where each sentence is encoded to a fixed-sized vector in a complete isolated manner and the two vectors for the corresponding sentences are used in predicting the degree of matching. The others are joint methods that allow to utilize interactive features like attentive information between the sentences. 

In the former paradigm, because two sentences have no interaction, they can not utilize interactive information during the encoding procedure. In our work, we adopted a joint method which enables capturing interactive information for performance improvements. 
Furthermore, we employ a substantially deeper recurrent network for sentence matching like deep neural machine translator (NMT) \cite{wu2016gnmt}. Deep recurrent models are more advantageous for learning long sequences and outperform the shallower architectures. However, the attention mechanism is unstable in deeper models with the well-known vanishing gradient problem. Though GNMT \cite{wu2016gnmt} uses residual connection between recurrent layers to allow better information and gradient flow, there are some limitations. The recurrent hidden or attentive features are not preserved intact through residual connection because the summation operation may impede the information flow in deep networks. 

Inspired by Densenet \cite{huang2017densely}, we propose a densely-connected recurrent network where the recurrent hidden features are retained to the uppermost layer. In addition, instead of the conventional summation operation, the concatenation operation is used in combination with the attention mechanism to preserve co-attentive information better. 
The proposed architecture shown in Figure \ref{fig:arch} is called DRCN which is an abbreviation for \textit{Densely-connected Recurrent and Co-attentive neural Network}. 
The proposed DRCN can utilize the increased representational power of deeper recurrent networks and attentive information. 
Furthermore, to alleviate the problem of an ever-increasing feature vector size due to concatenation operations, we adopted an autoencoder and forwarded a fixed length vector to the higher layer recurrent module as shown in the figure.
\sh{DRCN is, to our best knowledge, the first generalized version of DenseRNN which is expandable to deeper layers with the property of controllable feature sizes by the use of an autoencoder.}

We evaluate our model on three sentence matching tasks: \textit{natural language inference}, \textit{paraphrase identification} and \textit{answer sentence selection}. Experimental results on five highly competitive benchmark datasets (SNLI, MultiNLI, QUORA, TrecQA and SelQA) show that our model significantly outperforms the current state-of-the-art results on 
\shh{most of} the tasks.


\section{Related Work}
\label{sec:rel}
Earlier approaches of sentence matching mainly relied on conventional methods such as syntactic features, transformations or relation extraction \cite{romano2006investigating,wang2007jeopardy}.  These are restrictive in that they work only on very specific tasks.

The developments of large-scale annotated datasets \cite{snliemnlp2015,williams2017broad} and deep learning algorithms have led a big progress on matching natural language sentences. Furthermore, the well-established attention mechanisms endowed richer information for sentence matching by providing alignment and dependency relationship between two sentences.
The release of the large-scale datasets also has encouraged the developments of the learning-centered approaches to semantic representation. The first type of these approaches is sentence-encoding-based methods~\cite{conneau2017supervised,choi2017learning,nie2017shortcut,shen2018reinforced} where sentences are encoded into their own sentence representation without any cross-interaction. Then, a classifier such as a neural network is applied to decide the relationship based on these independent sentence representations. These sentence-encoding-based methods are simple to extract sentence representation and are able to be used for transfer learning to other natural language tasks \cite{conneau2017supervised}. On the other hand, the joint methods, which make up for the lack of interaction in the former methods, use cross-features as an attention mechanism to express the word- or phrase-level alignments for performance improvements~\cite{wang2017bilateral,chen2017enhanced,gong2018natural,yang2016anmm}.

Recently, the architectural developments using deeper layers have led more progress in performance. The residual connection is widely and commonly used to increase the depth of a network stably \cite{he2016deep,wu2016gnmt}.
More recently, Huang \etal \cite{huang2017densely} enable the features to be connected from lower to upper layers using the concatenation operation without any loss of information on lower-layer features.

External resources are also used for sentence matching. Chen \etal \cite{chen2017natural,chen2017enhanced} used syntactic parse trees or \sh{lexical databases like} WordNet to measure the semantic relationship among the words and Pavlick \etal \cite{pavlick2015adding} added interpretable semantics to the paraphrase database.

Unlike these, in this paper, we do not use any \sh{such} external resources.
Our work belongs to the joint approaches 
which uses densely-connected recurrent and co-attentive information to enhance representation power for semantic sentence matching.

\section{Methods}
\label{sec:method}
In this section, we describe our sentence matching architecture DRCN which is composed of the following three components: (1) word representation layer, (2) attentively connected RNN and (3) interaction and prediction layer. We denote two input sentences as $P = \{p_{1}, p_{2}, \cdots, p_{I}\}$ and $Q = \{q_{1}, q_{2}, \cdots, q_{J}\}$ where $p_i$/$q_j$ is the $i^{th}$/$j^{th}$ word of the sentence $P$/$Q$ and $I$/$J$ is the word length of $P$/$Q$. 
The overall architecture of the proposed DRCN is shown in Fig. \ref{fig:arch}.

\subsection{Word Representation Layer}

To construct the word representation layer, we concatenate word embedding, character representation and the exact matched flag which was used in \cite{gong2018natural}.

In word embedding, each word is represented as a $d$-dimensional vector by using a pre-trained word embedding method such as GloVe \cite{pennington2014glove} or Word2vec \cite{mikolov2013distributed}. In our model, a word embedding vector can be updated or fixed during training. The strategy whether to make the pre-trained word embedding be trainable or not is heavily task-dependent. Trainable word embeddings capture the characteristics of the training data well but can result in overfitting. On the other hand, fixed (non-trainable) word embeddings lack flexibility on task-specific data, while it can be robust for overfitting, especially for less frequent words. We use both the trainable embedding $e_{p_i}^{tr}$ and the fixed (non-trainable) embedding $e_{p_i}^{fix}$ to let them play complementary roles in enhancing the performance of our model. This technique of mixing trainable and non-trainable word embeddings is simple but yet effective. 

The character representation $c_{p_i}$ is calculated by feeding randomly initialized character embeddings into a convolutional neural network with the max-pooling operation. The character embeddings and convolutional weights are jointly learned during training. 

Like \cite{gong2018natural}, the exact match flag $f_{p_i}$ is activated if the
same word is found in the other sentence. 

Our final word representational feature $p_i^w$ for the word $p_i$ is composed of four components as follows:
\begin{equation}
\label{eq:word}
\begin{split}
e_{p_i}^{tr} = E^{tr}(p_i),& \quad
e_{p_i}^{fix} = E^{fix}(p_i) \\
c_{p_i} = \text{Ch}&\text{ar-Conv}(p_i) \\
p_i^w = [e_{p_i}^{tr}; & e_{p_i}^{fix}; c_{p_i}; f_{p_i}].
\end{split}
\end{equation}
Here, $E^{tr}$ and $E^{fix}$ are the trainable and non-trainable (fixed) word embeddings respectively. Char-Conv is the character-level convolutional operation and $[\cdot \ ; \ \cdot]$ is the concatenation operator. For each word in both sentences, the same above procedure is used to extract word features.

\subsection{Densely connected Recurrent Networks}

The ordinal stacked RNNs (Recurrent Neural Networks) are composed of multiple RNN layers on top of each other, with the output sequence of previous layer forming the input sequence for the next. More concretely, let $H_l$ 
be the $l^{th}$ 
RNN layer in a stacked RNN.
Note that in our implementation, we employ the bidirectional LSTM \sh{(BiLSTM)}
as a base block of $H_l$. At the time step $t$, an ordinal stacked RNN is expressed as follows:
\begin{equation}
\begin{split}
h_t^l =& H_l(x_t^{l}, h_{t-1}^l) \\
x_t^l &= h_t^{l-1}.
\end{split}
\label{eq:stackedrnn}
\end{equation}
%
While this architecture enables us to build up higher level representation, deeper networks have difficulties in training due to the exploding or vanishing gradient problem.

To encourage gradient to flow in the backward pass, residual connection \cite{he2016deep} is introduced which bypasses the non-linear transformations with an identity mapping. Incorporating this into (\ref{eq:stackedrnn}), it becomes
\begin{equation}
\begin{split}
h_t^l &= H_l(x_t^{l}, h_{t-1}^l) \\
x_t^l &= h_t^{l-1}+x_t^{l-1}.
\end{split}
\end{equation}

However, the summation operation in the residual connection may impede the information flow in the network \cite{huang2017densely}. Motivated by Densenet \cite{huang2017densely}, we employ direct connections using the concatenation operation from any layer to all the subsequent layers so that the features of previous layers are not to be modified but to be retained as they are as depicted in Figure \ref{fig:arch}.
The densely connected recurrent neural networks can be described as
\begin{equation}
\begin{split}
h_t^l &= H_l(x_t^{l}, h_{t-1}^l) \\
x_t^l &= [h_t^{l-1}; x_t^{l-1}]. 
\end{split}
\end{equation}
The concatenation operation enables the hidden features to be preserved until they reach to the uppermost layer and all the previous features work for prediction as collective knowledge \cite{huang2017densely}.

\subsection{Densely-connected Co-attentive networks}

Attention mechanism, which has largely succeeded in many domains \cite{wu2016gnmt,vaswani2017attention}, is a technique to learn effectively where a context vector is matched conditioned on a specific sequence.

Given two sentences, a context vector is calculated based on an attention mechanism focusing on the relevant part of the two sentences at each RNN layer. The calculated attentive information represents soft-alignment between two sentences. In this work, we also use an attention mechanism. 
We incorporate co-attentive information into densely connected recurrent features using the concatenation operation, so as not to lose any information (Fig. \ref{fig:arch}). This concatenated recurrent and co-attentive features which are obtained by  densely connecting the features from the undermost to the uppermost layers, enrich the collective knowledge for lexical and compositional semantics.

The attentive information $a_{p_i}$ of the $i^{th}$ word $p_i \in P$ against the sentence $Q$ is calculated as a weighted sum of $h_{q_j}$'s which are weighted by the softmax weights as follows :
\begin{equation}
\begin{split}
a_{p_i} &= \sum_{j=1}^{J} \alpha_{i, j} h_{q_j} \\
\alpha_{i, j} &= \frac{exp(e_{i, j})}{\sum_{k=1}^{J}exp(e_{i, k})} \\
e_{i, j} &= \cos(h_{p_i}, h_{q_j})
\end{split}
\label{eq:attention}
\end{equation}

Similar to the densely connected RNN hidden features, we concatenate the attentive context vector $a_{p_i}$ with triggered vector $h_{p_i}$ so as to retain attentive information as an input to the next layer: 
\begin{equation}
\begin{split}
h_t^l &= H_l(x_t^{l}, h_{t-1}^l) \\
x_t^l &= [h_t^{l-1}; a_t^{l-1}; x_t^{l-1}]. 
\end{split}
\end{equation}

\subsection{Bottleneck component}
Our network uses all layers' outputs as a community of semantic knowledge. However, this network is a structure with increasing input features as layers get deeper, and has a large number of parameters especially in the fully-connected layer. To address this issue, we employ an autoencoder as a bottleneck component. Autoencoder is a compression technique that reduces the number of features while retaining the original information, which can be used as a distilled semantic knowledge in our model. Furthermore, this component increased the test performance by working as a regularizer \sh{in our experiments}.

\subsection{Interaction and Prediction Layer}
To extract a proper representation for each sentence, we apply the step-wise max-pooling operation over densely connected recurrent and co-attentive features (pooling in Fig. \ref{fig:arch}). More specifically, if the output of the final RNN layer is a 100d vector for a sentence with 30 words, a $30\times 100$ matrix is obtained which is max-pooled column-wise such that the size of the resultant vector $p$ or $q$ is 100.  
Then, we aggregate these representations $p$ and $q$ for the two sentences $P$ and $Q$ in various ways in the interaction layer and the final feature vector $v$ for semantic sentence matching is obtained as follows:
\begin{equation}
v = [p; q; p+q; p-q; |p-q|].
\end{equation}
Here, \nj{the operations $+$, $-$ and $|\cdot|$ 
} 
are performed element-wise \nj{to infer the relationship between two sentences.} The element-wise subtraction $p-q$ is an asymmetric operator for one-way type tasks such as \textit{natural language inference} or \textit{answer sentence selection}. 

Finally, based on previously aggregated features $v$, we use two fully-connected layers with ReLU activation followed by one fully-connected output layer. Then, the softmax function is applied to obtain a probability distribution of each class.
The model is trained end-to-end by minimizing the multi-class cross entropy loss and the reconstruction loss of autoencoders.

\section{Experiments}
\label{sec:exp}


We evaluate our matching model on five popular and well-studied benchmark datasets for three challenging sentence matching tasks:  (i) SNLI and MultiNLI for natural language inference; (ii) Quora Question Pair for paraphrase identification; and (iii) TrecQA and SelQA for answer sentence selection in question answering. 
Additional details about the above datasets
can be found in the supplementary materials.

\begin{table}
\resizebox{\linewidth}{!}
{
\begin{tabular}{p{2.8in}}
	\hline
	\textbf{Premise} {\em two bicyclists in spandex and helmets in a race pedaling uphill.} \\
    \textbf{Hypothesis} {\em A pair of humans are riding their
    bicycle with tight clothing, competing with each other.} \\
    \textbf{Label}  {\em\{\textbf{entailment}; neutral; contradiction\}}
\\
	\hline
    \textbf{Premise} {\em Several men in front of a white building.} \\
    \textbf{Hypothesis} {\em Several people in front of a gray building.} \\
    \textbf{Label}  {\em\{entailment; neutral; \textbf{contradiction}\}}
\\
	\hline
\end{tabular}
}
\caption{Examples of \textit{natural language inference}.}
\label{tab:NLI}
\end{table}

\subsection{Implementation Details}
We initialized word embedding with 300d GloVe vectors pre-trained from the 840B Common Crawl corpus \cite{pennington2014glove}, while the word embeddings for the out-of-vocabulary words were initialized randomly. We also randomly initialized character embedding with a 16d vector and extracted 32d character representation with a convolutional network. For the densely-connected recurrent layers, we stacked 5 layers each of which have 100 hidden units. We set 1000 hidden units with respect to the fully-connected layers. The dropout was applied after the word and character embedding layers with a keep rate of 0.5. It was also applied before the fully-connected layers with a keep rate of 0.8. 
For the bottleneck component, we set 200 hidden units as encoded features of the autoencoder with a dropout rate of 0.2.
The batch normalization was applied on the fully-connected layers, only for the one-way type datasets. 
The RMSProp optimizer with an initial learning rate of 0.001 was applied. 
The learning rate was decreased by a factor of 0.85 when the dev accuracy does not improve. All weights except embedding matrices are constrained by L2 regularization with a regularization constant $\lambda =10^{-6}$. The sequence lengths of the sentence are all different for each dataset: 35 for SNLI, 55 for MultiNLI, 25 for Quora question pair and 50 for TrecQA. 
The learning parameters were selected based on the best performance on the dev set. 
We employed 8 different randomly initialized sets of parameters with the same model for our ensemble approach.

\subsection{Experimental Results}

\subsubsection{SNLI and MultiNLI}

We evaluated our model on the natural language inference task over SNLI and MultiNLI datasets. Table \ref{tab:exp_snli} shows the results on SNLI dataset of our model with other published models.
Among them, \shh{ESIM+ELMo and LM-Transformer} are the current state-of-the-art models. 
However, 
they use additional 
\sh{contextualized word representations} from language models 
\shh{as an externel knowledge.} 
The proposed DRCN obtains an accuracy of 88.9\% which is a competitive score although we do not use any external \shh{knowledge like ESIM+ELMo and LM-Transformer}. The ensemble model achieves an accuracy of 90.1\%, which sets the new state-of-the-art performance. 
\shh{Our ensemble model with 53m parameters (6.7m$\times$8) outperforms the LM-Transformer whose the number of parameters is 85m.}
Furthermore, in case of the encoding-based method, we obtain the best performance of 86.5\% without the co-attention and exact match flag. 

Table \ref{tab:exp_mnli} shows the results on \textsc{matched} and \textsc{mismatched} \nj{problems} of MultiNLI dataset. 
\shh{Our plain DRCN has a competitive performance without any contextualized knowledge. And, by combining DRCN with the ELMo, one of the contextualized embeddings from language models, our model outperforms the LM-Transformer which has 85m parameters with fewer parameters \shh{of 61m}. From this point of view, the combination of our model with a contextualized knowledge is a good option to enhance the performance.}

\begin{table}[t]
\centering
\resizebox{\linewidth}{!}
{
\begin{tabular}{lcc}
	\hline
   	\textbf{Models} & \textbf{Acc.} & \textbf{$|\theta|$}  \\
	\hline
    \multicolumn{3}{c}{\textit{\textbf{Sentence encoding-based method}}} \\
	\hline
	BiLSTM-Max \cite{conneau2017supervised} & 84.5 & 40m   \\
	Gumbel TreeLSTM \cite{choi2017learning} & 85.6 & 2.9m   \\
    CAFE \cite{tay2017compare} & 85.9 & 3.7m  \\
	Gumbel TreeLSTM \cite{choi2017learning} & 86.0 & 10m   \\
	Residual stacked \cite{nie2017shortcut} & 86.0 & 29m  \\ 
	Reinforced SAN \cite{shen2018reinforced} & 86.3 & 3.1m \\ 
	Distance SAN \cite{im2017distance} & 86.3 & 3.1m  \\          
	\textbf{DRCN} (- Attn, - Flag) & \textbf{86.5} & 5.6m  \\
	\hline
    \multicolumn{3}{c}{\textit{\textbf{Joint method (cross-features available)}}} \\
	\hline
    DIIN \cite{gong2018natural} & 88.0 / 88.9 & 4.4m \\
    ESIM \cite{chen2017enhanced} & 88.0 / 88.6 & 4.3m \\
    BCN+CoVe+Char \cite{mccann2017learned} & 88.1 / \ \ -\ \ \ \ \ & 22m \\
    DR-BiLSTM \cite{ghaeini2018dr} & 88.5 / 89.3 & 7.5m \\
    CAFE \cite{tay2017compare} & 88.5 / 89.3 & 4.7m \\
    KIM \cite{chen2017natural} & 88.6 / 89.1 & 4.3m  \\
    ESIM+ELMo \cite{Peters2018elmo} & 88.7 / 89.3 & 8.0m  \\
    \shh{LM-Transformer \cite{radford2018improving}} & \textbf{89.9} / \ \ -\ \ \ \ \ & 85m  \\
    \textbf{DRCN} (- AE) & 88.7 / \ \ -\ \ \ \ \ & 20m \\
    \textbf{DRCN} & 88.9 / \textbf{90.1} & 6.7m \\
	\hline
\end{tabular}
}
\caption{Classification accuracy (\%) for natural language inference on SNLI test set. $|\theta|$ denotes the number of parameters in each model.}
\label{tab:exp_snli}
\end{table}

\begin{table}[t]
\centering
\resizebox{\linewidth}{!}
{
\begin{tabular}{lcc}
	\hline
	\multirow{2}{*}{\textbf{Models}} & \multicolumn{2}{c}{\textbf{Accuracy (\%)}} \\
	 & \textbf{\textsc{matched}} & \textbf{\textsc{mismatched}} \\
	\hline
    ESIM \cite{williams2017broad} &	72.3 &	72.1 \\
	DIIN \cite{gong2018natural} &	78.8 & 77.8 \\
	CAFE \cite{tay2017compare} &	78.7 & 77.9 \\
	\shh{LM-Transformer \cite{radford2018improving}} & \textbf{82.1} &	\textbf{81.4} \\
	\textbf{DRCN} & 79.1 &	78.4 \\
  	\hline
	DIIN* \cite{gong2018natural} &	80.0 &	78.7 \\
	CAFE* \cite{tay2017compare} & 80.2	  &	 79.0 \\
	\textbf{DRCN}* &	\textbf{80.6}  &	\textbf{79.5} \\
	\shh{\textbf{DRCN+ELMo}*} &	\shh{\textbf{82.3}}  &\shh{\textbf{81.4}} \\
   	\hline
\end{tabular}
}
\caption{Classification accuracy for natural language inference on MultiNLI test set. * denotes ensemble methods.}
\label{tab:exp_mnli}
\end{table}

\begin{table}[t]
\centering
\resizebox{\linewidth}{!}
{
\begin{tabular}{lc}
	\hline
	\textbf{Models} & \textbf{Accuracy (\%)}  \\
	\hline
    Siamese-LSTM \cite{wang2017bilateral} & 82.58  \\
    MP LSTM \cite{wang2017bilateral} & 83.21  \\
    L.D.C. \cite{wang2017bilateral} & 85.55  \\
    BiMPM \cite{wang2017bilateral} & 88.17  \\
    pt-DecAttchar.c \cite{tomar2017neural} & 88.40  \\
    DIIN \cite{gong2018natural} & 89.06  \\
    \textbf{DRCN} & \textbf{90.15}  \\
	\hline
    DIIN* \cite{gong2018natural} & 89.84  \\
    \textbf{DRCN}* & \textbf{91.30}  \\
   	\hline
\end{tabular}
}
\caption{Classification accuracy for paraphrase identification on Quora question pair test set. * denotes ensemble methods.}
\label{tab:exp_quora}
\end{table}

\begin{table}[!]
\subfloat[TrecQA: \textit{raw} and \textit{clean}]{
\resizebox{\linewidth}{!}
{
\begin{tabular}{lcc}
	\hline
	 \textbf{Models}  & \textbf{MAP} & \textbf{MRR} \\
	\hline
    \multicolumn{3}{c}{\textit{\textbf{Raw version}}} \\
	\hline
	aNMM \cite{yang2016anmm} & 0.750 & 0.811  \\
    PWIM \cite{he2016pairwise} & 0.758 & 0.822  \\
	MP CNN \cite{he2015multi} &	0.762 &	0.830    \\
    HyperQA \cite{tay2017hyper} &	0.770 &	0.825  \\
	PR+CNN \cite{rao2016noise} &	0.780 &	0.834 \\
	\textbf{DRCN}  & \textbf{0.804} & \textbf{0.862}  \\
	\hline
    \multicolumn{3}{c}{\textit{\textbf{clean version}}} \\
	\hline
    HyperQA \cite{tay2017hyper} & 0.801 & 0.877 \\
    PR+CNN \cite{rao2016noise} &	0.801 &	0.877 \\
    BiMPM \cite{wang2017bilateral} & 0.802 & 0.875 \\				   
    Comp.-Aggr. \cite{bian2017compare} & 0.821 & 0.899 \\  
    IWAN \cite{shen2017inter} & 0.822 & 0.889  \\
    \textbf{DRCN} & \textbf{0.830} & \textbf{0.908} \\
	\hline    
\end{tabular}}}
\hfill
\subfloat[SelQA]{
\resizebox{\linewidth}{!}
{
\begin{tabular}{lcc}
	\hline
    \textbf{Models} & \textbf{MAP} & \textbf{MRR} \\
	\hline
	CNN-DAN \cite{santos2017learning}  & 0.866 & 0.873 \\
	CNN-hinge \cite{santos2017learning}  & 0.876 & 0.881 \\
    ACNN \cite{shen2017adaptive} &  0.874 & 0.880 \\
    AdaQA \cite{shen2017adaptive} &  0.891 & 0.898 \\    
    \textbf{DRCN}  & \textbf{0.925} & \textbf{0.930} \\
	\hline
\end{tabular}}}
\caption{Performance for answer sentence selection on TrecQA and selQA test set.}
\label{tab:exp_trec}
\end{table}

\subsubsection{Quora Question Pair}

Table \ref{tab:exp_quora} shows our results on the Quora question pair dataset. BiMPM using the multi-perspective matching technique between two sentences reports baseline performance of a L.D.C. network and basic multi-perspective models \cite{wang2017bilateral}. We obtained accuracies of 90.15\% and 91.30\% in single and ensemble methods, respectively, surpassing the previous state-of-the-art model of DIIN.

\subsubsection{TrecQA and SelQA}

Table \ref{tab:exp_trec} shows the performance of different models on
TrecQA and SelQA datasets for answer sentence selection task that aims to select a set of candidate answer sentences given a question. 
Most competitive models \cite{shen2017inter,bian2017compare,wang2017bilateral,shen2017adaptive} also use attention methods for words alignment between question and candidate answer sentences. However, the proposed DRCN using collective attentions over multiple layers, achieves the new state-of-the-art performance, exceeding the current state-of-the-art performance significantly on both datasets.

\begin{figure}[!t]
\centering
\resizebox{\linewidth}{!}
{\begin{tabular}{lc}
	\hline
    Models							 & \textbf{Accuracy (\%)} \\
	\hline
	\hline
    (1) \hspace{1mm} DRCN & 89.4 \\
	\hline   
    (2) \hspace{2mm} $-$ autoencoder & 89.1 \\
	\hline
    (3) \hspace{2mm} $-$ $E^{tr}$ & 88.7 \\
    (4) \hspace{2mm} $-$ $E^{fix}$ & 88.9 \\
	\hline
    (5) \hspace{2mm} $-$ dense(Attn.) & 88.7 \\
    (6) \hspace{2mm} $-$ dense(Rec.) & 88.8 \\
    (7) \hspace{2mm} $-$ dense(Rec. \& Attn.)  & 88.5 \\
    (8) \hspace{2mm} $-$ dense(Rec. \& Attn.)  & 
    \multirow{2}{*}{88.7} \\
        \hspace{6.3mm} $+$ res(Rec. \& Attn.) &  \\
    (9) \hspace{2mm} $-$ dense(Rec. \& Attn. \& Emb)  & 
	\multirow{2}{*}{88.4} \\
        \hspace{6.3mm} $+$ res(Rec. \& Attn.) &   \\
	\hline
    (10) \hspace{2mm} $-$ dense(Rec. \& Attn. \& Emb) & 87.8 \\
    (11) \hspace{2mm} $-$ dense(Rec. \& Attn. \& Emb) - Attn. & 85.3 \\
  	\hline
\end{tabular}}
\captionof{table}{Ablation study results on the SNLI dev sets.}
\label{tab:exp_abl}
\hfill
  \includegraphics[width=\linewidth]{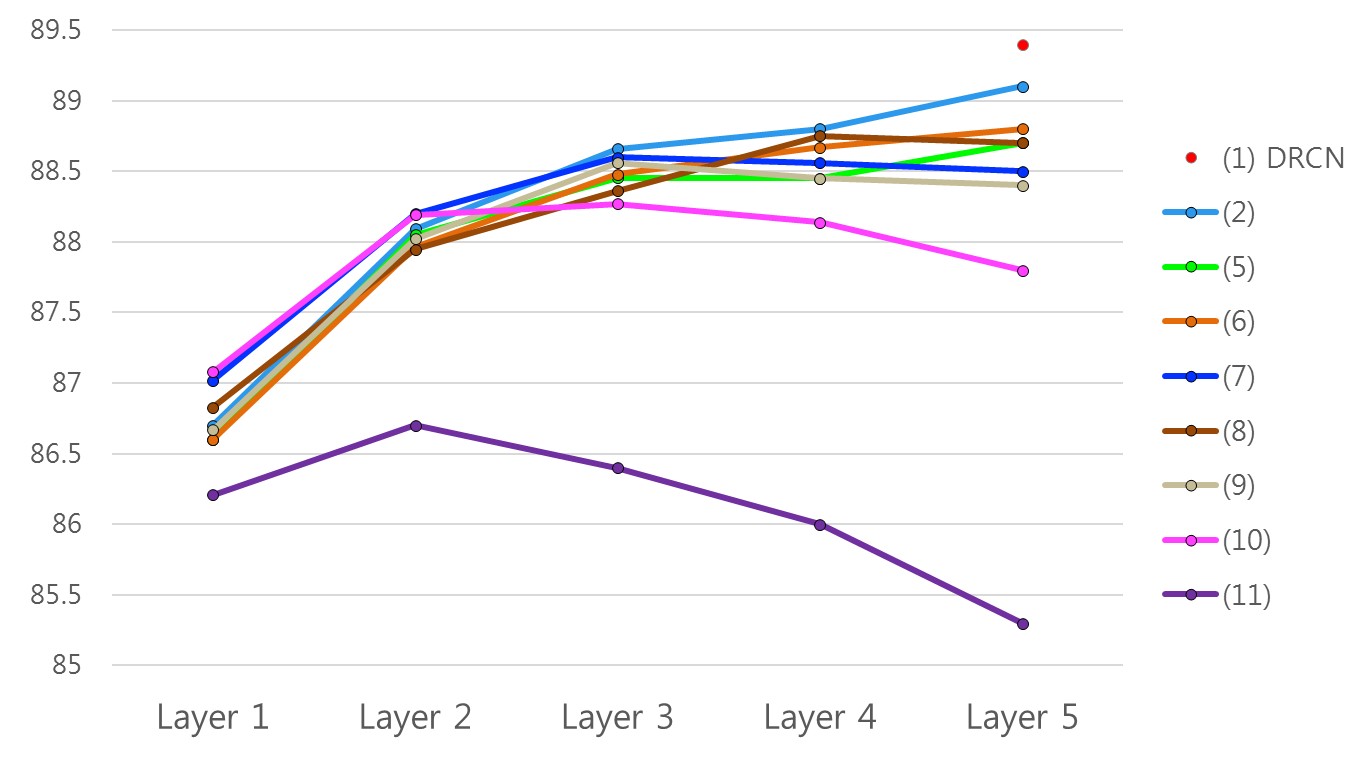}
  \captionof{figure}{Comparison of models on every layer in ablation study. (best viewed in color)}
  \label{fig:graph}

\end{figure} 

\subsection{Analysis}

\subsubsection{Ablation study}

We conducted an ablation study on the SNLI dev set as shown in Table \ref{tab:exp_abl}, where we aim to examine the effectiveness of our word embedding technique as well as 
the proposed densely-connected recurrent and co-attentive features. 
Firstly, we verified the effectiveness of the autoencoder as a bottleneck component in (2). 
Although the number of parameters in the DRCN significantly decreased as shown in Table \ref{tab:exp_snli}, we could see that the performance was rather higher because of the regularization effect. Secondly, we study how 
the technique of mixing trainable and fixed word embeddings
contributes to the performance in models (3-4). After removing $E^{tr}$ or $E^{fix}$ in eq. (\ref{eq:word}), the performance degraded, slightly. 
The trainable embedding $E^{tr}$ seems more effective than the fixed embedding $E^{fix}$.
Next, the effectiveness of dense connections was tested in models (5-9). In (5-6), we removed dense connections only over co-attentive or recurrent features, respectively. The result shows that the dense connections over attentive features are more effective. In (7), we removed dense connections over both co-attentive and recurrent features, and the performance degraded to 88.5\%. In (8), we replace dense connection with residual connection only over recurrent and co-attentive features. It means that only the word embedding features are densely connected to the uppermost layer while recurrent and attentive features are connected to the upper layer using the residual connection. In (9), we removed additional dense connection over word embedding features from (8). The results of (8-9) demonstrate that the dense connection using concatenation operation over deeper layers, has more powerful capability retaining collective knowledge to learn textual semantics. The model (10) is the basic 5-layer RNN with attention and (11) is the one without attention. The result of (10) shows that the connections among the layers are important to help gradient flow. And, the result of (11) shows that the attentive information functioning as a soft-alignment is significantly effective in semantic sentence matching.

\begin{table}[t!]
\centering
\scalebox{0.823}
{
\begin{tabular}{lcccc}
	\hline
    \textbf{Category} & \textbf{ESIM} & \textbf{DIIN} & \textbf{CAFE} & \textbf{DRCN}  \\
	\hline
	  \multicolumn{5}{c}{\textbf{Matched}} \\
	\hline
    Conditional	& \textbf{100}	&57&70&	65 \\
    Word overlap	& 50	&79&82	&\textbf{89} \\
    Negation	& 76	&78&76	&\textbf{80}	 \\
    Antonym& 67	&\textbf{82}&\textbf{82}&\textbf{82}  \\
    Long Sentence & 75	&81&79	&\textbf{83} \\
    Tense Difference & 73	&\textbf{84}&82&82 \\
    Active/Passive	& 88	&93&\textbf{100}	&87  \\
    Paraphrase	& 89	&88&88&	\textbf{92}  \\
    Quantity/Time & 33	&53&53	&\textbf{73} \\
    Coreference	& \textbf{83}	&77&80	&80	 \\
    Quantifier	& 69	&74&75	&\textbf{78}  \\
    Modal	& 78&\textbf{84}&81&81 \\
    Belief	& 65	&\textbf{77}& \textbf{77}&	76 \\
    \hline
    Mean	&72.8	&77.46&78.9	&\textbf{80.6}  \\
	Stddev	&16.6	&10.75&10.2	&\textbf{6.7} \\
	\hline
	\hline
    \multicolumn{5}{c}{\textbf{Mismatched}}  \\
	\hline
    Conditional	&	60&69	&85&	\textbf{89} \\
    Word overlap &62&\textbf{92}&	87	&89 \\
    Negation	&71	&77&\textbf{80}	&78 \\
    Antonym	 &58&\textbf{80}&\textbf{80}&\textbf{80} \\
    Long Sentence	&	69&73&	77	&\textbf{84} \\
    Tense Difference&79	&78&\textbf{89}&	83 \\
    Active/Passive&	91&70&	90	&\textbf{100} \\
    Paraphrase	&84	&\textbf{100}&95&90 \\
    Quantity/Time &54&69	&62	&\textbf{80} \\
    Coreference &75&79&	83&	\textbf{87} \\
    Quantifier	 &72&78&	80&	\textbf{82} \\
    Modal	&	76&	75&81	&\textbf{87} \\
    Belief &	67&81	&83	&\textbf{85} \\
    \hline
    Mean	&	70.6&78.53	&82.5&	\textbf{85.7} \\
	Stddev	&	10.2&8.55&	7.6&	\textbf{5.5} \\
	\hline
    \end{tabular}
}
\caption{Accuracy (\%) of Linguistic correctness on MultiNLI dev sets.}
\label{tab:linguistic_mnli}
\end{table}

The performances of models having different number of recurrent layers are also reported in Fig. \ref{fig:graph}. The models (5-9) which have connections between layers, are more robust to the increased depth of network, however, the performances of (10-11) tend to degrade as layers get deeper. In addition, the models with dense connections rather than residual connections, have higher performance in general. Figure \ref{fig:graph} shows that the connection between layers is essential, especially in deep models, endowing more representational power, and the dense connection is more effective than the residual connection.


\begin{figure*}[t]
  \centering  
  \scalebox{0.92}{
  \subfloat[entailment]{
  \label{subfig:entailment}
  \includegraphics[width=0.9\textwidth]{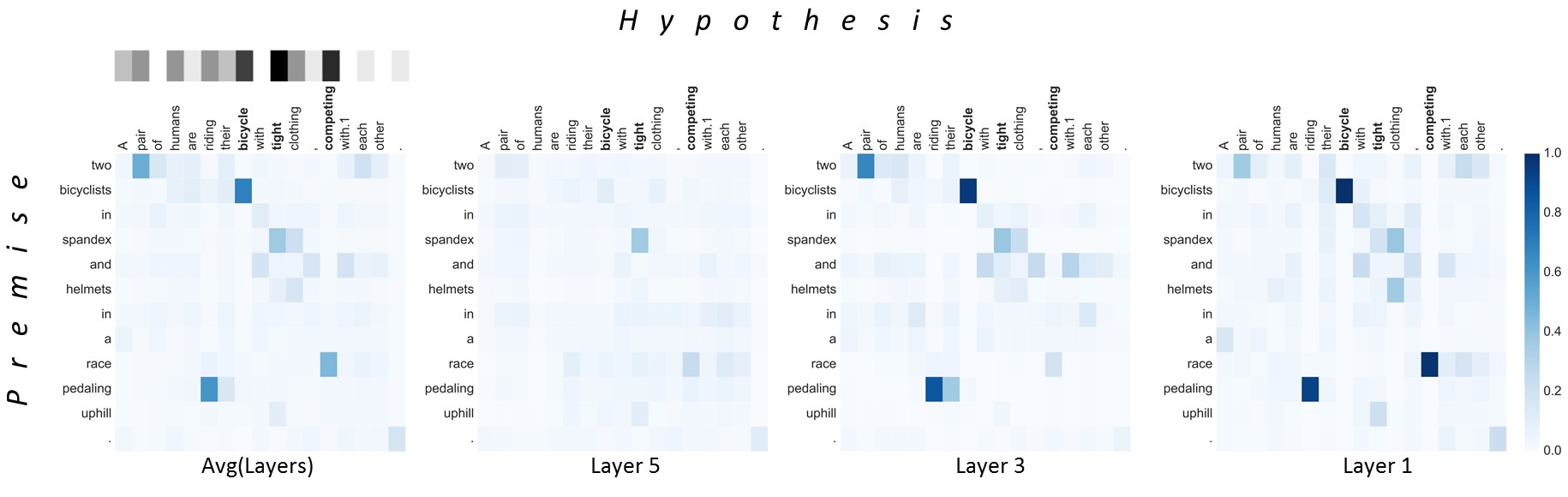}
  }}
  
  \scalebox{0.92}{
  \subfloat[contradiction]{
  \label{subfig:contradiction}
  \includegraphics[width=0.9\textwidth]{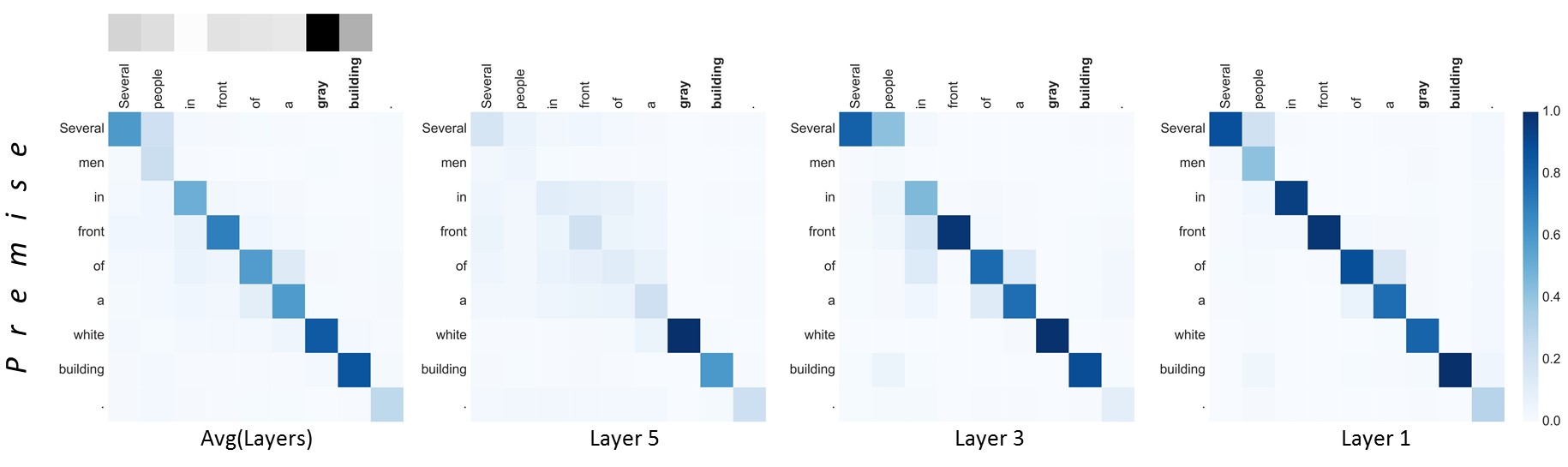}
  }}
  
  \caption{Visualization of attentive weights and the rate of max-pooled position. The darker, the higher. See supplementary materials for a comparison with other models \sh{that use the residual connections.}
}  
  \label{fig:exp_att}
\end{figure*}

\subsubsection{Word Alignment and Importance}

Our densely-connected recurrent and co-attentive features are connected to the classification layer through the max pooling operation such that all max-valued features of every layer affect the loss function and perform a kind of deep supervision \cite{huang2017densely}. Thus, we could cautiously interpret the classification results using our attentive weights and max-pooled positions. The attentive weights contain information on how two sentences are aligned and the numbers of max-pooled positions in each dimension play an important role in classification.

Figure \ref{fig:exp_att} shows the attention map ($\alpha_{i,j}$ in eq. (\ref{eq:attention})) on each layer of the samples in Table \ref{tab:NLI}. The Avg(Layers) is the average of attentive weights over 5 layers and the gray heatmap right above the Avg(Layers) is the rate of max-pooled positions. The darker indicates the higher importance in classification. In the figure, we can see that \textit{tight}, \textit{competing} and \textit{bicycle} are more important words than others in classifying the label. The word \textit{tight clothing} in the hypothesis can be inferred from \textit{spandex} in the premise. And \textit{competing} is also inferred from \textit{race}. 
Other than that, the \textit{riding} is matched with \textit{pedaling}, and \textit{pair} is matched with \textit{two}. Judging by the matched terms, the model is undoubtedly able to classify the label as an entailment, correctly.

In Figure \ref{fig:exp_att} (b), most of words in both the premise and the hypothesis coexist except \textit{white} and \textit{gray}. In attention map of layer 1, the same or similar words in each sentence have a high correspondence (\textit{gray} and \textit{white} are not exactly matched but have a linguistic relevance). However, as the layers get deeper, the relevance between \textit{white building} and \textit{gray building} is only maintained as a clue of classification (See layer 5). Because \textit{white} is clearly different from \textit{gray}, our model determines the label as a contradiction.

The densely connected recurrent and co-attentive features are well-semanticized over multiple layers as collective knowledge. And the max pooling operation selects the soft-positions that may extract the clues on inference correctly.

\subsubsection{\sh{Linguistic Error Analysis}}
\sh{We conducted a linguistic error analysis on MultiNLI, and compared DRCN with the ESIM, DIIN and CAFE. We used annotated subset provided by the MultiNLI dataset, and each sample belongs to one of the 13 linguistic categories. The results in table \ref{tab:linguistic_mnli} show that our model generally has a good performance than others on most categories. Especially, we can see that ours outperforms much better on the Quantity/Time category which is one of the most difficult problems. Furthermore, \nj{our DRCN shows the highest mean and the lowest stddev for both \textsc{matched} and \textsc{mismatched} problems, which indicates that it not only results in a competitive performance but also has a consistent performance.}}

\section{Conclusion}
\label{sec:conclusion}
In this paper, we introduce a densely-connected recurrent and co-attentive network (DRCN) for semantic sentence matching. We connect the recurrent and co-attentive features from the bottom to the top layer without any deformation. These intact features over multiple layers compose a community of semantic knowledge and outperform the previous deep RNN models using residual connections. In doing so, bottleneck components are inserted to reduce the size of the network. 
\sh{Our proposed model is the first generalized version of DenseRNN which can be expanded to deeper layers with the property of controllable feature sizes by the use of an autoencoder.}
We additionally show the interpretability of our model using the attentive weights and the rate of max-pooled positions. Our model achieves the state-of-the-art performance on 
\shh{most of the}
datasets of three highly challenging natural language tasks. 
\shh{Our proposed method using the collective semantic knowledge is expected to be applied to the various other natural language tasks.}

\bibliographystyle{aaai}
\small
\bibliography{aaai2019}

\section{Supplementary Material}

\subsection{Datasets}

\noindent \textbf{A. SNLI} is a collection of 570k human written sentence pairs based on image captioning, supporting the task of natural language inference \cite{snliemnlp2015}. The labels are composed of entailment, neutral and contradiction. The data splits are provided \nj{in} \cite{snliemnlp2015}. 

\noindent \textbf{B. MultiNLI,} also known as Multi-Genre NLI, has 433k sentence pairs whose size and mode of collection are modeled closely like SNLI. MultiNLI offers ten distinct genres (FACE-TO-FACE, TELEPHONE, 9/11, TRAVEL, LETTERS, OUP, SLATE, VERBATIM, GOVERNMENT and FICTION) of written and spoken English data. Also, there are matched dev/test sets which are derived from the same sources as those in the training set, and mismatched sets which do not closely resemble any seen at training time. The data splits are provided \nj{in} \cite{williams2017broad}.

\noindent  \textbf{C. Quora Question Pair} consists of over 400k question pairs based on actual \texttt{quora.com} questions. Each pair contains a binary value indicating whether the two questions are paraphrase or not. The training-dev-test splits for this dataset are provided \nj{in} \cite{wang2017bilateral}.

\noindent \textbf{D. TrecQA} provided \nj{in} \cite{wang2007jeopardy} was collected from TREC Question Answering tracks 8-13. There are two versions of data due to different pre-processing methods, namely clean and raw \cite{rao2016noise}. We evaluate our model on both data and follow the same data split as provided \nj{in} \cite{wang2007jeopardy}. We use official evaluation metrics of MAP (Mean Average Precision) and MRR (Mean Reciprocal Rank), which are standard metrics in information retrieval and question answering tasks.

\noindent \textbf{E. SelQA} consists of questions generated through crowdsourcing and the answer senteces are extracted from the ten most prevalent topics (Arts, Country, Food, Historical Events, Movies, Music, Science, Sports, Travel and TV) in the English Wikipedia. We also use MAP and MRR for our evaluation metrics, and the data splits are provided in \cite{jurczyk2016selqa}.

\newpage
\subsection{Visualization on the comparable models}

\label{sec:supplemental}
We study how the attentive weights flow as layers get deeper in each model using the dense or residual connection. We used the samples of the SNLI dev set in Table 1.

\begin{figure*}[t]
  \centering
  \includegraphics[width=\linewidth]{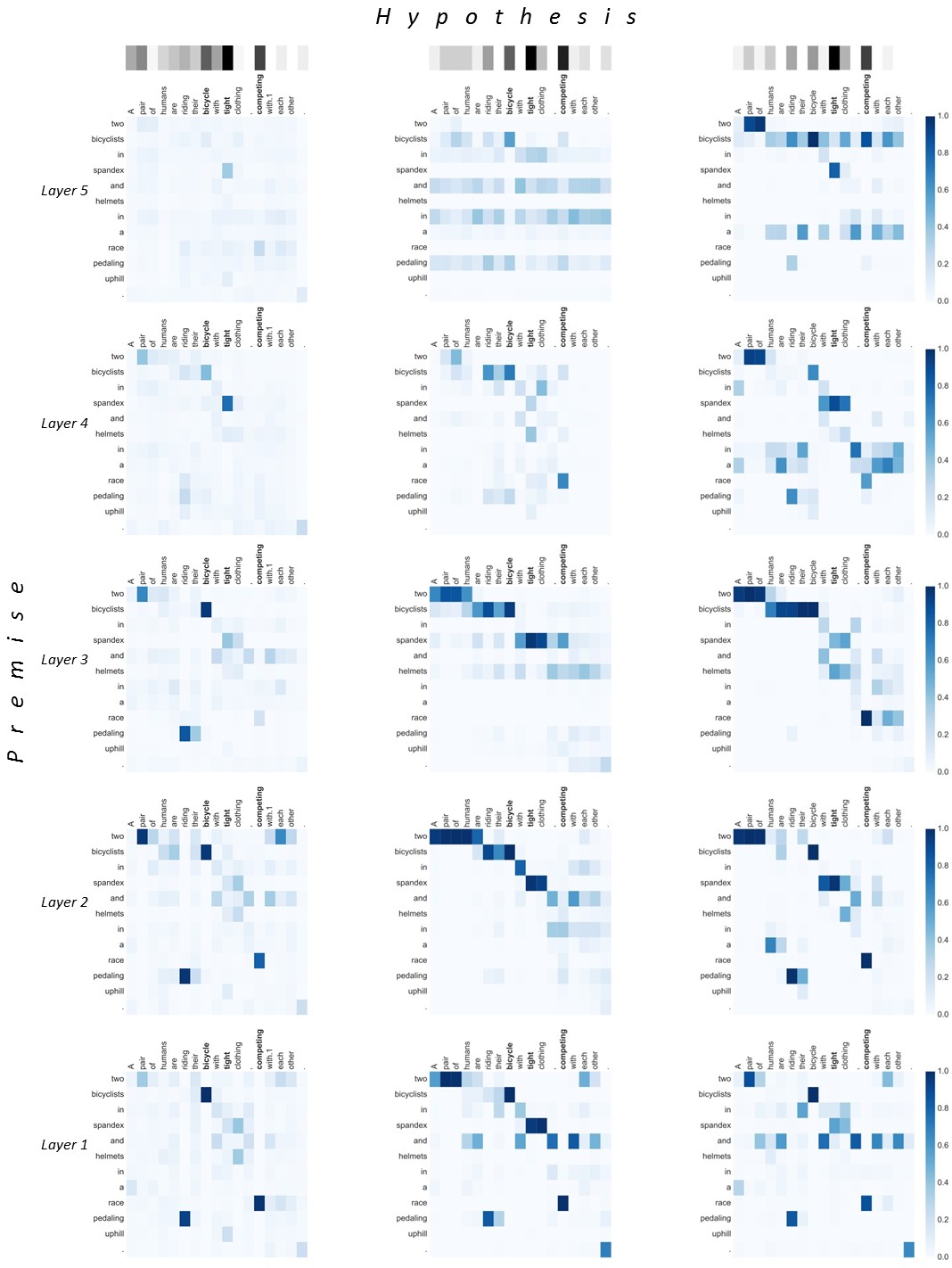}
  \caption{Visualization of attentive weights on the \textit{entailment} example. The premise is ``\textit{two bicyclists in spandex and helmets in a race pedaling uphill.}" and the hypothesis is ``\textit{A pair of humans are riding their
    bicycle with tight clothing, competing with each other.}". The attentive weights of DRCN, Res1, and Res2 are presented from left to right.}
  \label{fig:en}
\end{figure*}

\begin{figure*}[t]
  \centering
  \includegraphics[width=\linewidth]{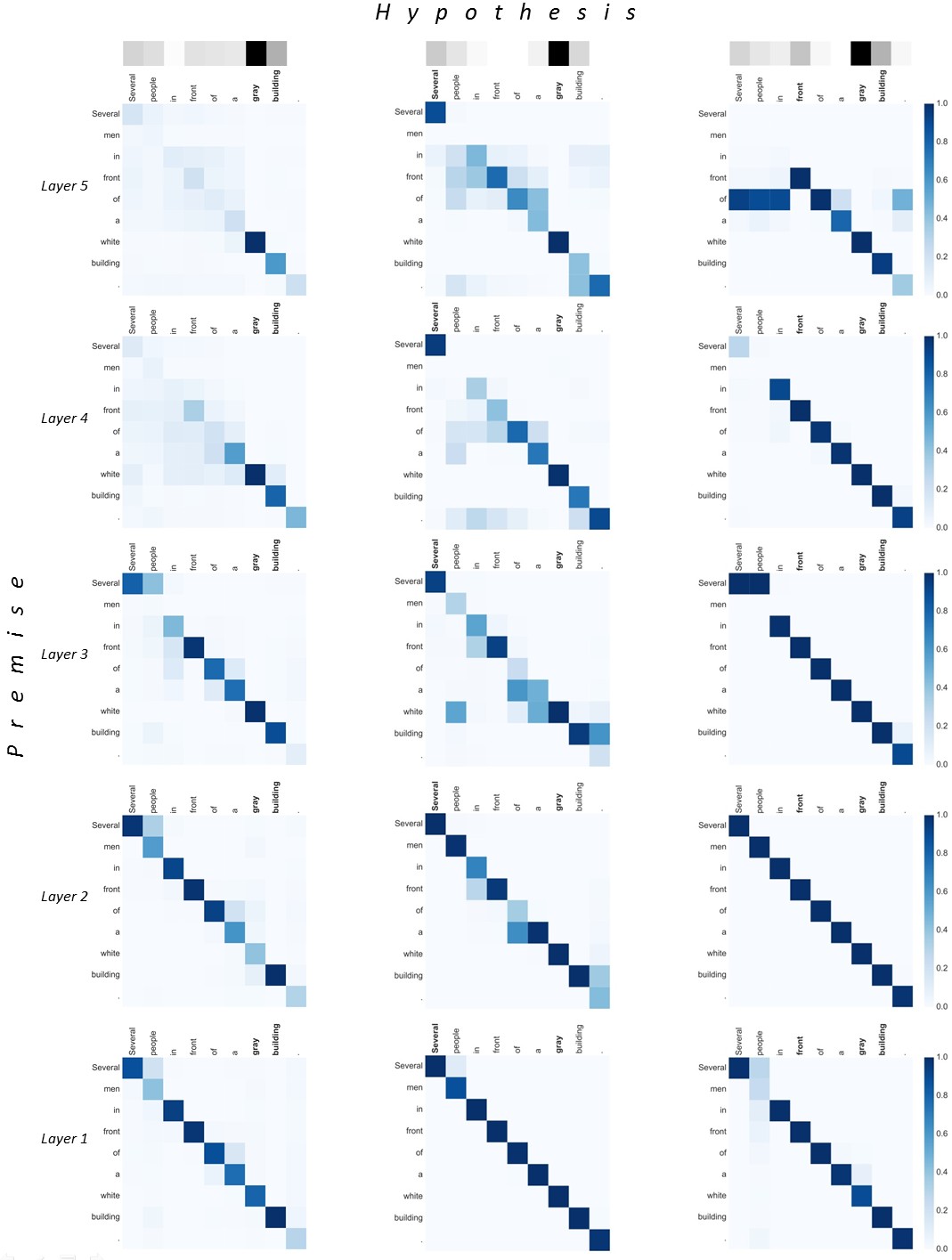}
  \caption{Visualization of attentive weights on the \textit{contradiction} example. The premise is ``\textit{Several men in front of a white building.}"  and the hypothesis is ``\textit{Several people in front of a gray building.}". The attentive weights of DRCN, Res1, and Res2 are presented from left to right.}
  \label{fig:con}
\end{figure*}

Figure \ref{fig:en} and \ref{fig:con} show the attention map on each layer of the models of DRCN, Table 6 (8), and Table 6 (9). In the model of Table 6 (8), we replaced the dense connection with the residual connection only over recurrent and co-attentive features. And, in the model of Table 6 (9), we removed additional dense connection over word embedding features from Table 6 (8). We denote the model of Table 6 (9) as Res1 and the model of Table 6 (8) as Res2 for convenience. 

In Figure \ref{fig:en}, DRCN does not try to find the right alignments at the upper layer if it already finds the rationale for the prediction at the relatively lower layer. This is expected that the DRCN use the features of all the preceding layers as a collective knowledge. While Res1 and Res2 have to find correct alignments at the top layer, however, there are some misalignments such as \textit{competing} and \textit{bicyclists} rather than \textit{competing} and \textit{race} in Res2 model.

In the second example in Figure \ref{fig:con}, although the DRCN couldn't find the clues at the lower layer, it gradually finds the alignments, which can be a rationale for the prediction. At the 5th layer of DRCN, the attentive weights of \textit{gray building} and \textit{white building} are significantly higher than others. On the other hand, the attentive weights are spread in several positions in both Res1 and Res2 which use residual connection.

\end{document}